\documentclass[10pt,twocolumn,letterpaper]{article}

\usepackage{cvpr}              

\usepackage{algpseudocode} 
\usepackage{algorithm}
\usepackage{algorithmicx}

\usepackage{stfloats}
\usepackage{subcaption}
\renewcommand{\paragraph}[1]{\vspace{.5em}\noindent\textbf{#1.}}

\definecolor{cvprblue}{rgb}{0.21,0.49,0.74}
\usepackage[pagebackref,breaklinks,colorlinks,allcolors=cvprblue]{hyperref}
\usepackage{svg}
\usepackage{graphicx}
\usepackage{hyperref}
\usepackage{url}
\usepackage{amsmath} 
\usepackage{amssymb}
\usepackage{multirow}
\usepackage{graphicx}
\usepackage{booktabs}
\usepackage{array}
\usepackage{bm}
\usepackage{dsfont}
\usepackage{wrapfig}
\usepackage{makecell}
\usepackage{float} 
\usepackage[accsupp]{axessibility}  

\def\paperID{} 
\def\confName{CVPR}
\def\confYear{2026}

\title{Noise-Aware Few-Shot Learning through Bi-directional Multi-View \\Prompt Alignment}

\author{Lu Niu$^{1,2}$,\hspace{2pt} Cheng Xue$^{1,2}$\thanks{Corresponding author.}\\
$^{1}$Southeast University\hspace{5pt} $^{2}$AIIA, Ministry of Education, China\\
{\tt\small \{lniu,cxue\}@seu.edu.cn}
}

\begin{document}
\maketitle
\begin{abstract}
Vision-language models offer strong few-shot capability through prompt tuning but remain vulnerable to noisy labels, which can corrupt prompts and degrade cross-modal alignment. Existing approaches struggle because they often lack the ability to model fine-grained semantic cues and to adaptively separate clean from noisy signals. To address these challenges, we propose NA-MVP, a framework for \textbf{N}oise-\textbf{A}ware few-shot learning through bi-directional \textbf{M}ulti-\textbf{V}iew \textbf{P}rompt alignment. NA-MVP is built upon a key conceptual shift: \emph{robust prompt learning requires moving from global matching to region-aware alignment that explicitly distinguishes clean cues from noisy ones}. To realize this, NA-MVP employs (1) multi-view prompts combined with unbalanced optimal transport to achieve fine-grained patch-to-prompt correspondence while suppressing unreliable regions; (2) a bi-directional prompt design that captures complementary clean-oriented and noise-aware cues, enabling the model to focus on stable semantics; and (3) an alignment-guided selective refinement strategy that uses optimal transport to correct only mislabeled samples while retaining reliable data. Experiments on synthetic and real-world noisy benchmarks demonstrate that NA-MVP consistently outperforms state-of-the-art baselines, confirming its effectiveness in enabling robust few-shot learning under noisy supervision.
\end{abstract}

\section{Introduction}
\label{sec:intro}
Vision–language models (VLMs), such as CLIP~\cite{radford2021learning}, have advanced multimodal understanding by embedding images and text into a shared semantic space. Building on this, prompt learning adapts VLMs to downstream tasks by optimizing a small set of learnable textual embeddings while keeping the backbone frozen~\cite{zhou2022learning, zhou2022conditional}. This paradigm is especially appealing in few-shot and resource-limited settings due to its parameter efficiency, modularity, and fast adaptation. However, real-world deployments frequently face noisy supervision, and the few-shot regime exacerbates this vulnerability: with only a handful of examples per class, even a small number of corrupted labels can disproportionately bias gradient updates and induce spurious correlations.

\begin{figure}[t]
    \centering
    \includegraphics[width=8cm]{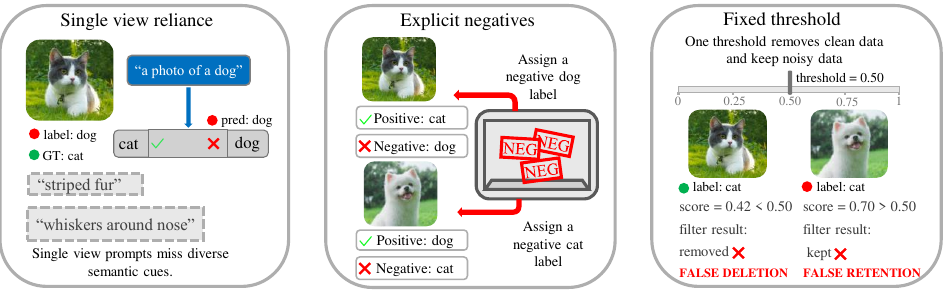}
    \caption{Limitations of existing prompt learning approaches under noisy labels. \textbf{Single-view reliance}: Limited prompts miss diverse visual patterns. \textbf{Explicit negatives}: Fixed negatives impose rigid supervision. \textbf{Fixed threshold}: Coarse denoising lets noise propagate.}
    \label{fig:limitations}
\end{figure}

Recent studies suggest that prompt learning can be made robust to label noise~\cite{wu2023prompt}, inspiring combinations with noisy-label learning such as negative learning~\cite{sun2022dualcoop,wei2024vision} and noisy-label selection~\cite{guo2024joapr,pan2025nlprompt}. Yet, as summarized in Figure~\ref{fig:limitations}, they still face \textit{key limitations}. First, prompt expressiveness is constrained, since most methods employ only one or two learnable prompts (a positive and a negative pair)~\citep{wu2023prompt,wei2024vision}, enforcing a single-view alignment that cannot capture diverse and fine-grained cues that are essential for reducing the influence of noisy labels in few-shot settings. Second, assigning an explicit negative label to each image imposes a rigid supervision signal tied to a fixed counter class, and such hard negatives are often inaccurate or uninformative, making the optimization process less reliable in noisy settings. Third, denoising is typically coarse, relying either on fixed confidence thresholds or pseudo-labeling without 
selective correction, leading to error propagation. These limitations highlight a missing perspective in prior work: \emph{robust noisy few-shot learning requires adaptively decomposing and aligning clean and noisy semantics at a fine-grained, region-aware level}. 

To address these challenges, we propose NA-MVP, a framework for noisy few-shot learning. NA-MVP combines multi-view, fine-grained patch-to-prompt alignment with Unbalanced Optimal Transport (UOT), allowing local features to be partially matched with multiple prompt views and mitigating the limitations of single-view prompting. To avoid rigid
negative supervision, we introduce a bi-directional prompt design that jointly learns clean-oriented and noise-aware prompts, where the noise-aware view serves as an implicit negative and provides more flexible guidance under noisy labels. Finally, a prompt-guided selective refinement module uses alignment signals to identify unreliable
samples and correct them via classical OT, offering a more targeted alternative to confidence-based relabeling. Our main contributions are summarized as follows:
\begin{itemize}
    \item \textbf{A new conceptual perspective for few shot learning with noisy label.} We introduce a new formulation of robustness in prompt learning: robust noisy few-shot learning requires decomposing and aligning clean and noisy semantics in a \emph{region-aware, class-conditional} manner, moving beyond global image–prompt matching adopted in prior work.
    \item \textbf{Bi-directional multi-view prompts for noise-aware alignment.} We design clean-oriented and noise-aware prompt views to capture reliable and corrupted cues respectively. Coupled with unbalanced patch-to-prompt alignment, this enables the model to downweight noisy regions and enhance consistent semantic signals.

    \item \textbf{Selective label refinement guided by alignment signals.} We develop a prompt-guided selective refinement mechanism that uses bi-directional alignment cues to identify mislabeled samples and correct them via classical OT, avoiding the over-correction issues of global pseudo-labeling approaches.

    \item We validate NA-MVP on multiple benchmarks and noise settings, showing consistent gains and robustness under noisy supervision.
\end{itemize}

\section{Related Work}
\label{sec:formatting}
\subsection{Learning with noisy labels.}
Learning with noisy labels (LNL) presents a significant challenge in training models that generalize well without overfitting to noisy labels. Existing approaches include robust loss functions~\citep{zhang2018generalized,wang2019symmetric,ma2020normalized,wei2023mitigating}, loss correction~\citep{chang2017active,arazo2019unsupervised,xia2019anchor}, robust noise regularization~\citep{wei2020combating,wei2021open,iscen2022learning,ko2023gift}, and sample selection~\citep{xue2019robust,xue2020cascaded,kim2021fine,karim2022unicon,xue2022robust,feng2023ot,li2023disc,xu2023usdnl,wei2024vision,xue2025dier}. Sample selection methods often rely on the small-loss criterion, which may discard clean hard samples and retain noisy ones. Label correction  strategies, like MLC~\citep{zheng2021meta} and SELC~\citep{lu2022selc}, aim to correct noisy annotations by generating pseudo-labels from model predictions. However, these methods typically process each sample independently, overlooking the relationships between data points, which can lead to suboptimal corrections. To better exploit global structure, recent works~\citep{xia2022ot,feng2023ot,chang2023csot} based on OT align feature distributions for improving pseudo-labeling. Recently,  prompt learning has shown promise in noisy settings~\citep{wu2023prompt,wei2024vision,pan2025nlprompt}. However, existing prompt-based LNL approaches still inherit core limitations of traditional methods: they typically operate at the global image–prompt level and rely on single-view or explicit negative cues, making them insensitive to fine-grained inconsistencies between clean and corrupted signals. As a result, their robustness can degrade significantly
in few-shot regimes where noisy labels disproportionately influence prompt semantics.

\subsection{Prompt Learning in Vision-Language Models.}
Prompt learning, initially developed in natural language processing, has become a central technique for adapting vision–language models (VLMs)~\citep{jia2021scaling,radford2021learning,yu2022coca}. While early models such as CLIP relied on manually crafted prompts, recent work focuses on learning continuous prompt embeddings. CoOp~\citep{zhou2022learning} introduces learnable prompts in the continuous space, and CoCoOp~\citep{zhou2022conditional} further adapts them at the image level to improve generalization to unseen classes. This paradigm has inspired a broad line of extensions~\citep{shu2022test,derakhshani2023bayesian,khattak2023maple,
khattak2023self,liu2023hierarchical,roy2023consistency,zhu2023prompt}. However, using a single prompt~\citep{zhou2022learning} limits the ability to capture diverse visual cues, motivating multi-prompt designs~\citep{lu2022prompt,sun2022dualcoop}. CLIPN~\citep{wang2023clipn}
employs a positive/negative prompt pair for OOD detection, where the negative prompts serve as class-agnostic cues to identify distribution shift. PLOT~\citep{chen2022plot} aligns multiple prompts with local image features through optimal transport to enhance image–text correspondence. While effective, these methods are developed for clean data or OOD generalization and do not address the challenges posed by noisy labels and extremely limited supervision in the noisy few-shot setting. Inspired by these works, we propose a framework that combines bi-directional and multi-view prompt learning to better distinguish clean and noisy semantics.

\subsection{Optimal Transport.}
OT provides a principled way to compare probability distributions by finding the most efficient mapping between them at a given cost. It defines the Wasserstein distance~\citep{peyre2019computational} and has been widely adopted in machine learning and computer vision. However,  the high computational complexity of OT was a bottleneck until Cuturi introduced entropic regularization, enabling efficient computation via the Sinkhorn algorithm~\citep{distances2013lightspeed}. To improve flexibility, UOT~\citep{lombardi2015eulerian,chizat2018unbalanced} was introduced, replacing the strict mass conservation constraint in classical OT with soft penalization terms~\citep{frogner2015learning,lahn2023combinatorial}. These advances have enabled OT to support a wide range of applications, including semi-supervised learning~\citep{taherkhani2020transporting,lai2022sar,wang2022solar}, object detection~\citep{ge2021ota,yang2021rethinking,de2023unbalanced}, generative models~\citep{balaji2020robust,daniels2021score,choi2023generative}, domain adaptation~\citep{chang2022unified,wang2024outlier}, learning with noisy labels~\citep{feng2023ot,chang2023csot} and others. Building on these advances, our method adopts UOT with relaxed mass constraints to align local image features with multi-view prompts, allowing the model to focus on reliable features while suppressing noise. Meanwhile, classical OT with strict mass preservation is used to refine noisy labels by aligning global image features with class-level prompts, ensuring reliable label correction. This design leverages the complementary strengths of both OT variants for robust learning under label noise.

\begin{figure*}[t]
    \centering
    \includegraphics[width=17.5cm]{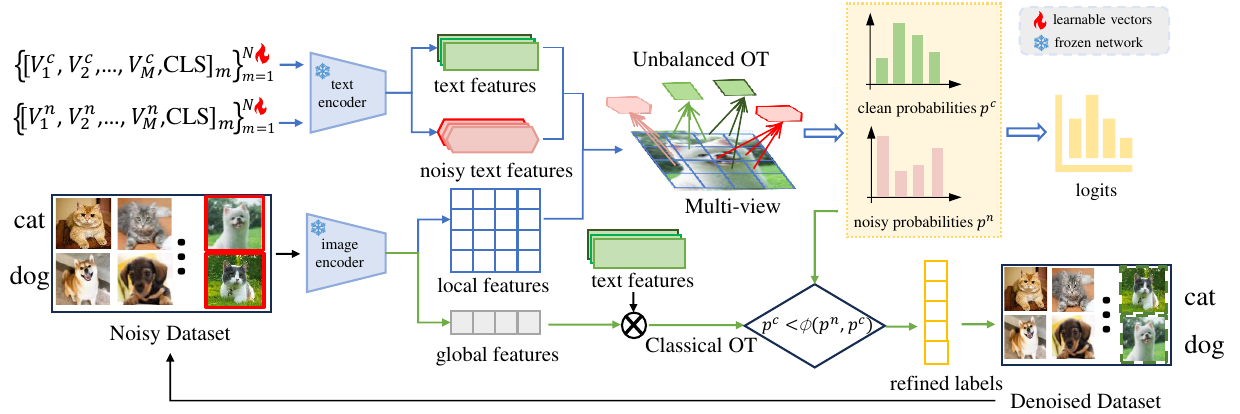}
    \caption{
    Overview of the NA-MVP framework. Our framework consists of two key modules: (1) Noise-aware alignment (blue arrows): Multiple clean and noise-aware prompts per class are encoded and aligned with local image patches via UOT to generate clean/noisy probabilities. (2) Selective label refinement (green arrows): An adaptive threshold $\phi$ derived from these probabilities identifies mislabeled samples, which are refined via classical OT by aligning global image features with clean text features. The two modules work together to iteratively update the training set while optimizing the prompts, producing a denoised dataset for robust prediction under noisy supervision.
    }
    \label{fig:framework}

\end{figure*}
\section{Methodology}
\noindent\textbf{Conceptual motivation.} Existing prompt-based LNL largely rely on global image–prompt coupling or explicit negative labels. No sample-dependent mechanisms are capable of separating clean signals from corrupted ones within an image. In contrast, NA-MVP is built on the idea that robustness should arise from fine-grained, region-aware alignment and sample-dependent label correction.

Our method consists of two key components: Bi-directional multi-view prompts for noise-aware alignment, where objects are observed from multiple perspectives. Selective noisy label refinement with OT, where the refinement is guided by prompts, as shown in Figure~\ref{fig:framework}. We present multi-view and bi-directional as a unified prompt design, since the two are tightly coupled in how they provide complementary semantics and generate alignment signals for refinement.

\paragraph{Problem definition}
Let \( \mathcal{D}_{\text{noisy}} = \{ (x_i, y_i) \}_{i=1}^D \) denote the noisy training set with images \( x_i \) and labels \( y_i \) from \( C \) classes. However, whether the given label is accurate or not is unknown. We classify the correctly labeled instances as clean, and the mislabeled ones as noisy. The goal of LNL is to train a model that maintains high test accuracy while minimizing the influence of label noise.

\subsection{Bi-directional Multi-view Prompts for Noise-aware Alignment}
To improve robustness under noisy supervision, we propose a bi-directional multi-view prompt learning strategy that integrates multiple clean-oriented and noise-aware prompts for each class. Unlike explicit negative learning~\citep{wei2024vision}, which relies on explicitly labeled negative classes, our method treats all unlabeled classes as implicit negatives, avoiding the need for additional annotations. Clean-oriented prompts focus on capturing class-relevant semantics, while noise-aware prompts function as adaptive filters to identify and suppress misleading or noisy signals. These two types of prompts are jointly optimized to facilitate noise-aware representation learning and enhance the model’s discrimination between clean and corrupted samples. This bi-directional prompt mechanism offers complementary perspectives on the data and serves as the foundation for downstream label refinement and denoising.

\paragraph{Prompt construction and feature encoding}
 For each class \text{k}, we construct two sets of learnable prompts: clean-oriented prompts \( \{\text{Prompt}_{m,k}^{c}\}_{m=1}^{N} \) and noise-aware prompts \( \{\text{Prompt}_{m,k}^{n}\}_{m=1}^{N} \), where \(N\) denotes the number of prompts for each class. Each prompt consists of M context tokens followed by a class-specific token:
\begin{align}
\text{Prompt}_{m,k}^{c} &= [V_1^{c}, V_2^{c}, \ldots, V_M^{c}, \texttt{CLS}_k] \\
\text{Prompt}_{m,k}^{n} &= [V_1^{n}, V_2^{n}, \ldots, V_M^{n}, \texttt{CLS}_k]
\end{align}
where \( V_i^{c} \) and \( V_i^{n} \) are learnable word embeddings, and \(\texttt{CLS}_k\) denotes the class-specific token. Clean and noisy prompts are encoded by text encoder to generate prompt feature sets \( \boldsymbol{G}_k^c \in\mathbb{R}^{N\times d}\) and \( \boldsymbol{G}_k^n\in\mathbb{R}^{N\times d} \), where d is the feature dimension. Given an input image \( \boldsymbol{x}_i \), the image encoder extracts a global feature \( \boldsymbol{f}_i \in \mathbb{R}^d \) and a local feature map \( \boldsymbol{F}_i = \{ f_l \}_{l=1}^L \in \mathbb{R}^{L\times d} \), where \( L = H \times W \), with \( H \) and \( W \) denoting the height and width of the feature map. 

\paragraph{Fine-grained noise-aware alignment}
To align local image features with multi-view prompts under noisy conditions, we employ UOT for noise-aware multi-view alignment. Unlike classical OT, UOT relaxes the strict mass conservation constraint,  allowing partial alignment and reducing overfitting to noise. This enables more selective matching between informative visual features and semantic prompts, preserving prompt diversity and improving alignment robustness.

Specifically, we treat \(\boldsymbol{F}_i \) and \( \boldsymbol{G}_k \)  as discrete distributions and compute the cost matrix \( \boldsymbol{C}_k = 1 - \boldsymbol{F}_i \boldsymbol{G}_k^\top \in \mathbb{R}^{L \times N}\) by using the cosine similarity. The UOT problem is then formulated as:
\begin{gather}
\label{eq:UOT}
d_{\text{UOT}}(k) = \min_{T \in \Pi(\mu, \nu)} \langle \boldsymbol{C}_k, T \rangle \\
\quad \Pi(\mu, \nu) = \left\{ T \in \mathbb{R}^{L \times N}_+ \mid T \mathds{1}_N \leq \mu, T^\top \mathds{1}_L = \nu \right\}
\end{gather}
where $\langle \cdot, \cdot \rangle$ is the Frobenius dot-product, \( \mu \in \mathbb{R}^L \) and \( \nu \in \mathbb{R}^N \) represent marginal probability vectors satisfying $\|\mu\|_1 \geq \|\nu\|_1 = \theta$, with $\theta \in [0,1]$ controlling the degree of partial matching. The relaxed constraints in UOT allow for partial alignment between image features and prompt embeddings, rather than enforcing full correspondence. This is particularly helpful under label noise, where forcing all features to align may introduce irrelevant or incorrect information. To efficiently solve the UOT problem, we apply the Sinkhorn distance with entropic regularization:
\begin{equation}
d_{\text{UOT}}(k) = \min_{T \in \Pi(\mu, \nu)} \langle \boldsymbol{C}_k, T \rangle + \epsilon \langle T, \log T \rangle,
\end{equation}
where $\epsilon \geq 0$ is a  hyper-parameter. This formulation can be interpreted as a Kullback-Leibler (KL) projection~\citep{benamou2015iterative}:
\begin{gather}
\label{eq:6}
d_{\text{UOT}}(k) = \min_{T \in \Pi(\mu, \nu)} \epsilon \text{ KL}(T \mid e^{-\boldsymbol{C}_k/\epsilon})
\end{gather}

To solve Eq.~\ref{eq:6}, we use a fast implementation of Dykstra's algorithm, which scales the iterative KL projection between \( \mathcal{C}_1 \) and \( \mathcal{C}_2 \) using matrix-vector multiplications. The optimization proceeds iteratively, initializing \( Q = \exp(-\boldsymbol{C}_k/\epsilon) \) and \( \nu^{(0)} = \mathds{1}_N \), and iteratively update the scaling vectors \( \mu^{(t)}\) and \( \nu^{(t)}\). After a few iterations, the transport plan $T^*$ is computed as:
\begin{equation}
T^* = \text{diag}(
\mu^{(t)}) Q \text{diag}(\nu^{(t)})
\end{equation}
where \( t \) is the iteration number. 
Once the $T^*$ is obtained, we use it to compute the UOT distance and optimize learnable vectors in the bi-directional multi-view prompts \( \{\text{Prompt}_{m}^{c/n}\}_{m=1}^{N} \). Overall, by relaxing the strict mass conservation constraint, UOT enables partial and adaptive alignment between local features and multi-view prompts. This noise-aware alignment allows the model to focus on semantically reliable regions while suppressing noisy or irrelevant patches, thereby preserving prompt diversity and improving robustness under label noise.

\paragraph{Image-text bi-directional prompt loss}
We further stabilize this process with an auxiliary image-text bi-directional prompt (ITBP) Loss. The design of ITBP loss follows the bi-directional contrastive loss introduced in CLIPN~\citep{wang2023clipn}, but its role  in our framework is fundamentally different. While CLIPN leverages this loss for out-of-distribution detection, ITBP in our setting is employed to explicitly separate clean and noisy semantics.

Specifically, we distinguish two types of interactions between an image and noise-aware prompts and encode them using a binary indicator $m$. Based on this indicator, the ITBP loss is given by:
\begin{equation}
\begin{split}
\mathcal{L}_{\text{itbp}} =-\frac{1}{N} \sum_{i=1}^{N} (1 - m_{ii}) \log(1 - p_{ii}^{\text{n}}) - \\ \frac{1}{N(N-1)} \sum_{i=1}^{N} \sum_{j \neq i} m_{ij} \log(p_{ij}^{\text{n}})
\end{split}
\end{equation}

Concretely, it encourages
image features to align more closely with clean prompts while being pushed away from corresponding noisy prompts and unrelated negatives. This objective reinforces the effectiveness of the bi-directional multi-view prompt design under noisy label supervision. 
\subsection{Selective label refinement with OT}\label{4.1}
\paragraph{Adaptive noise identification via bi-directional prompt alignment} To adaptively identify noisy labels, we measure the alignment between the local image feature \( \boldsymbol{F}_i \) and the clean prompt \( \boldsymbol{G}_k^{c} \) as well as the noise-aware prompt \( \boldsymbol{G}_k^{n} \). These similarities are computed using the UOT-based distance, as defined in Eq.~\ref{eq:UOT}, yielding similarities \( s_{i,k}^{c} \) and \( s_{i,k}^{n} \):
\begin{align}
\label{eq:sim}
s_{i,k}^c &= 1 - d_{\text{UOT}}(\boldsymbol{F}_i,\boldsymbol{G}_k^c ), \\
s_{i,k}^n &= 1 - d_{\text{UOT}}(\boldsymbol{F}_i, \boldsymbol{G}_k^n),
\end{align}
where higher values indicate stronger semantic consistency with the corresponding prompts. To obtain a normalized measure of the clean alignment, we convert the clean similarity into a probabilistic confidence:
\begin{equation}
p_{ik}^c =
\frac{\exp(s_{i,k}^c / \tau)}
{\sum_{j=1}^{C} \exp(s_{i,j}^c / \tau)},
\end{equation}
where \( \tau \) is a learnable temperature parameter. This probability reflects how strongly sample  \( \boldsymbol{x}_i \) aligns with the clean semantics of class k. Next, we derive an adaptive threshold \( \phi_{i,k} \) based on the relative alignment between the clean and noise-aware prompts:
\begin{equation}
\phi_{i,k} = \frac{\exp(s_{i,k}^n / \tau)}{\exp(s_{i,k}^c / \tau) + \exp(s_{i,k}^n / \tau)},
\end{equation}
Then, a sample is classified as clean if its clean-prompt confidence exceeds this adaptive threshold:
\begin{equation}
\mathcal{D}_{\text{clean}} = \left\{ (x_i, y_i) \mid p_{ik}^c > \phi_{i,k}, k = y_i \right\}.
\end{equation}

Samples that do not meet this condition are considered noisy. The empirical motivation behind this thresholding strategy is further discussed in Appendix. Rather than discarding these noisy samples, we refine their labels using a pseudo-labeling strategy based on classical OT. In this way, the bi-directional prompt framework not only identifies noisy samples but also re-integrates them with corrected supervision. The clean and noisy prompts are jointly optimized with the model parameters, enabling continuous refinement and better adaptation to the evolving feature space.

\paragraph{Label refinement via classical OT}
Classical OT is a powerful tool for aligning distributions, and it has been effectively used for generating pseudo-labels by matching samples to class distributions while preserving the global structure of the sample distribution through equality constraints. Here, we adopt classical OT to refine noisy labels in the context of noisy label learning. 

Given a noisy training set \( \mathcal{D}_{\text{noisy}} = \{(x_i, y_i)\}_{i=1}^D \), we extract global image features \( \boldsymbol{F} \in \mathbb{R}^{D \times d} \), and the prompt features \( \boldsymbol{G} \in \mathbb{R}^{C \times d} \) using a pre-trained vision-language model. Next, we compute the similarity matrix between the image and prompt features, \( \boldsymbol{F} \cdot \boldsymbol{G}^{\top}\), and use its negative logarithm as the cost matrix. To ensure proper alignment, we enforce uniform marginal distributions for both the samples and the classes. The OT problem is then formulated as:
\begin{gather}
d_{\text{OT}}(\mu, \nu) =\min_{\boldsymbol{T} \in \Pi(\boldsymbol{\mu}, \boldsymbol{\nu})} \langle  -\log ( \boldsymbol{F}\cdot \boldsymbol{G}^{\top}), \boldsymbol{T} \rangle
\\
\Pi(\mu, \nu) = \left\{ \mathbf{T} \in \mathbb{R}_+^{C\times D} \;\middle|\; \mathbf{T} \mathds{1}_D = \mu, \; \mathbf{T}^\top \mathds{1}_C = \nu \right\}
\end{gather}
where \( \mathds{1}_C \) is the vector of ones with length \( C \), representing the total probability mass of the noisy label distribution, and \( \mathds{1}_D \) is the vector of ones with length \( D \), representing the total probability mass of the sample distribution. These constraints ensure that the total probability mass is conserved across both the samples and the labels. Once the optimal transport plan \( T^* \) is computed, the pseudo-label for each image \( x_i \) is obtained by selecting the class with the highest transport mass:
\begin{equation}
\tilde{y}_i = \arg \max_j T^*_{ij}
\end{equation}

This process generates refined labels by using the transport plan \( T^* \) to assign the most probable class for each image. To further improve reliability, we integrate the adaptive threshold \( \phi_{i,k} \) defined earlier to identify potentially mislabeled samples. Only those samples whose similarity to clean prompts falls below the threshold are considered for refinement, ensuring that clean samples remain unaltered while noisy instances are corrected:
\begin{equation}
\mathcal{D}_{\text{refinement}} = \left\{ (x_i,\tilde{y}_i) \mid p_{ik}^c < \phi_{i,k}, k = y_i \right\}.
\end{equation}

The selective mechanism, built on the bi-directional multi-view prompt learning framework, enables the model to isolate and correct corrupted labels effectively. This not only improves the label quality but also enhances the robustness of the model under noisy supervision. Ultimately, the denoised training set is constructed by combining reliable clean samples with the refined noisy ones:
\begin{equation}
\mathcal{D}_{\text{denoised}} = \mathcal{D}_{\text{clean}} \cup \mathcal{D}_{\text{refinement}}.
\end{equation}

By training on \( \mathcal{D}_{\text{denoised}} \), the model benefits from both trustworthy clean supervision and corrected noisy labels, leading to more stable convergence and improved generalization performance.

\subsection{Training Details}

\paragraph{Training schedule} To improve robustness, we delay the label refinement process and only start modifying labels after $T_{\text{sup}}$ epochs. Details of the full training procedure are provided in Appendix. In the early phase, the model is trained on the noisy dataset with the Generalized Cross-Entropy (GCE) loss~\citep{zhang2018generalized} combined with the ITBP loss:
\begin{equation}
\mathcal{L}_{\text{sup}} = \mathcal{L}_{\text{gce}} + \lambda_{\text{i}} \cdot \mathcal{L}_{\text{itbp}},
\end{equation}
where \( \lambda_{\text{i}} \) controls the strength of auxiliary supervision. Once the refinement process is activated, noisy samples identified by our prompt-guided mechanism are selectively corrected, and training continues on the updated dataset with GCE loss. This delayed refinement allows the model to first acquire stable representations before adapting to cleaner supervision. A detailed sensitivity study on the effect of \( \lambda_{\text{i}} \) is reported in the Appendix.

\paragraph{Inference}
During inference, both clean and noise-aware prompt alignments are incorporated into the prediction. We first compute the noise-aware confidence for class k as:
\begin{gather}
p_{ik}^n = \frac{\exp(s_{i,k}^n / \tau)}{\exp(s_{i,k}^c/ \tau) + \exp(s_{i,k}^n/ \tau)}
\end{gather}

Using both the clean-prompt confidence \( p_{ik}^c \)  and the noise-aware confidence \( p_{ik}^n \) , the final probability of assigning label $k$ to image $x_i$ is defined as:
\begin{gather}
p(y = k \mid x_i) = (1 - p_{ik}^n) \cdot p_{ik}^c 
\end{gather}

 \begin{table*}[t]
  \caption{Comparison of methods under symmetric and asymmetric noise on five datasets. (\%)}
  \label{tab:noise-results}
  \centering
  \setlength{\tabcolsep}{8.5pt}
  \label{tab:1}
  \begin{tabular}{
    >{\centering\arraybackslash}m{2.2cm}  
    >{\centering\arraybackslash}m{2.2cm}  
    *{6}{c}                               
    *{2}{c}                               
  }
    \toprule
    \multirow{2}{*}{\textbf{Dataset}} & \multirow{2}{*}{\textbf{Method}} 
    & \multicolumn{6}{c}{\textbf{Noise rate: Sym}} 
    & \multicolumn{2}{c}{\textbf{Noise rate: Asym}} \\
    & & 0.125 & 0.25 & 0.375 & 0.5 & 0.625 & 0.75 & 0.25 & 0.5 \\
    \midrule
    \multirow{5}{*}{Caltech101}
    & CoOp~\citep{zhou2022learning}     & 86.43 & 81.03 & 76.73 & 70.90 & 61.33 & 46.90 & 75.23 & 49.43 \\
    & GCE~\citep{wu2023prompt}     & 92.00 & 90.90 & 90.80 & 89.30 & 86.70 & 79.03 & 91.20 & 85.80 \\
    & JoAPR~\citep{guo2024joapr}   & 88.32 & 87.85 & 87.00 & 87.03 & 84.55 & 80.15 & 82.79 & 69.02 \\
    & NLPrompt~\citep{pan2025nlprompt} & 91.73 & 91.13 & 90.77 & 89.93 & 88.30 & 86.70 & 91.17 & 89.27 \\
    & NA-MVP    &  \textbf{92.07}  & \textbf{92.10} & \textbf{91.60} &   \textbf{91.30} & \textbf{90.07}  & \textbf{89.37} & \textbf{91.47} & \textbf{89.53} \\
    \midrule
    \multirow{5}{*}{DTD}
    & CoOp~\citep{zhou2022learning}    & 56.00 & 49.57 & 43.30 & 34.37 & 27.83 & 17.27 & 47.75 & 29.63 \\
    & GCE~\citep{wu2023prompt}     & 61.00 & 59.83 & 56.80 & 50.73 & 43.60 & 33.67 & 57.57 & 43.97 \\
    & JoAPR~\citep{guo2024joapr}    & 55.02 & 53.95 & 51.57 & 49.12 & 44.24 & 35.90 & 49.23 & 38.33 \\
    & NLPrompt~\citep{pan2025nlprompt} & 62.97 & 61.23 & 59.17 & 55.17 & 49.03 & 39.80 & 60.60 & 50.80 \\
    & NA-MVP    &  \textbf{63.73}  & \textbf{63.13} & \textbf{61.63} &   \textbf{58.50} & \textbf{52.93}  & \textbf{48.63} & \textbf{62.33} & \textbf{52.10}  \\
    \midrule
    \multirow{5}{*}{Flowers102}
    & CoOp~\citep{zhou2022learning}     & 88.93 & 83.50 & 77.93 & 70.10 & 55.60 & 37.17 & 74.70 & 42.60 \\
    & GCE~\citep{wu2023prompt}    & 88.80 & 88.33 & 86.73 & 84.07 & 78.37 & 70.37 & 86.37 & 69.93 \\
    & JoAPR~\citep{guo2024joapr}  & 84.90 & 84.70 & 79.75 & 77.13 & 69.65 & 64.20 & 79.57 & 55.47 \\
    & NLPrompt~\citep{pan2025nlprompt} & 93.87 & 92.57 & \textbf{92.73} & 89.90 & 84.77 & \textbf{76.80} & \textbf{93.40} & \textbf{81.10} \\
    & NA-MVP    &  \textbf{94.20}  & \textbf{93.30} & 92.00 &   \textbf{90.47} & \textbf{85.07}  & 76.47 & 91.37 & 78.43  \\
    \midrule
    \multirow{5}{*}{OxfordPets}
    & CoOp~\citep{zhou2022learning}    & 76.50 & 66.73 & 60.33 & 47.03 & 35.77 & 24.60 & 66.20 & 38.73 \\
    & GCE~\citep{wu2023prompt}     & 85.63 & 84.60 & 83.67 & 79.23 & 71.40 & 53.17 & 83.03 & 68.07 \\
    & JoAPR~\citep{guo2024joapr}   & 83.17 & 82.05 & 80.62 & 79.05 & 73.72 & 60.97 & 76.82 & 62.85 \\
    & NLPrompt~\citep{pan2025nlprompt} & 86.17 & 86.00 & 85.33 & 83.17 & 80.03 & 70.77 & 84.97 & 77.53 \\
    & NA-MVP    &  \textbf{88.50}  & \textbf{88.40} & \textbf{88.23} &   \textbf{88.13} & \textbf{86.93}  & \textbf{86.23} & \textbf{87.53} & \textbf{79.33}  \\
    \midrule
    \multirow{5}{*}{UCF101}
    & CoOp~\citep{zhou2022learning}     & 69.03 & 63.40 & 58.23 & 49.73 & 40.83 & 26.30 & 58.07 & 34.43 \\
    & GCE~\citep{wu2023prompt}   & 74.00 & 73.63 & 72.57 & 69.37 & 66.00 & 57.07 & 71.87 & 67.97 \\
    & JoAPR~\citep{guo2024joapr}   & 70.80 & 69.22 & 68.15 & 64.80 & 61.82 & 57.52 & 63.98 & 49.67 \\
    & NLPrompt~\citep{pan2025nlprompt} & 74.83 & 73.40 &  \textbf{72.83} & 70.33 & 68.10 & 60.53 & \textbf{73.58} & \textbf{65.97} \\
    & NA-MVP    &  \textbf{75.33}  & \textbf{74.03} & 72.30 &   \textbf{70.93} & \textbf{68.43}  & \textbf{63.93} & 73.40 & 65.40  \\
    \bottomrule
  \end{tabular}
\end{table*}

This bi-directional multi-view strategy encourages the model to prioritize clean samples while adaptively down-weighting potential noisy ones, leading to more stable and accurate learning under noisy supervision.

 \section{Experiments}
\label{sec5}
\subsection{Experimental Settings}
\paragraph{Datasets}
To evaluate the robustness of our method under label noise, we conduct experiments on five widely-used synthetic noisy datasets: Caltech101~\citep{fei2004learning}, DTD~\citep{cimpoi2014describing}, Flowers102~\citep{nilsback2008automated}, OxfordPets~\citep{parkhi2012cats}, and UCF101~\citep{peng2018two}. Following~\citep{pan2025nlprompt}, we utilize two types of label noise: symmetric noise (Sym) and asymmetric noise (Asym). For symmetric noise, labels in the training set are randomly flipped to other classes with equal probability; for asymmetric noise, labels are flipped to semantically similar neighboring classes.
In addition, we also evaluate on a real-world noisy dataset, Food101N~\citep{lee2018cleannet}, which inherently contains label noise without requiring synthetic corruption. A detailed introduction of each dataset can be found in the Appendix.

\paragraph{Implementation details}
We follow the experimental protocol of CoOp~\citep{zhou2022learning} for fair comparison. The model is optimized using SGD with an initial learning rate of 0.002, a momentum of 0.9 and a weight decay of 5×\(10^{-4}\). Unless otherwise specified, ResNet-50 is used as the image encoder and a 63M-parameter Transformer as the text encoder. For each dataset, we adopt a 16-shot training setup and evaluate on the full clean test set. We train for 50 epochs and use 16 shared context tokens appended to the class token. All reported results are averaged over three runs with different random seeds, and the best accuracy is highlighted in bold. All experiments are conducted on a single NVIDIA GeForce RTX 4090 GPU. More implementation details can be found in the Appendix.
\subsection{Performance Comparison}
We compare our method with strong baselines, including CoOp~\citep{zhou2022learning}, CoOp+GCE~\citep{wu2023prompt}, JoAPR~\citep{guo2024joapr} and NLPrompt~\citep{pan2025nlprompt}. Among them, CoOp serves as a standard prompt learning baseline, while CoOp+GCE, JoAPR and NLPrompt are specifically designed to address label noise in the few-shot prompt learning scenario. These comparisons allow us to comprehensively assess the effectiveness and robustness of our method in both synthetic and real-world noise settings. As shown in Table \ref{tab:noise-results}, our proposed NA-MVP consistently achieves the best overall performance across all five datasets. 
Furthermore, in scenarios with high levels of noise, our method exhibits a substantial performance advantage, highlighting its robustness in severely corrupted environments. These results demonstrate that NA-MVP effectively mitigates the impact of label noise and generalizes well under challenging settings. 

We further evaluate our method on the 
real-world noisy dataset Food101N. As shown in Table~\ref{tab:food}, NA-MVP consistently outperforms NLPrompt under all few-shot settings. Notably, as the number of shots decreases, 
NLPrompt's performance drops significantly, while NA-MVP remains stable and robust. These results highlight the effectiveness of NA-MVP in few-shot learning with real-world noise, underscoring its practicality for deployment.
\subsection{Ablation Studies and More Analysis}
\label{sec4.3}
We conduct a comprehensive ablation study to quantify the contributions of each core component in our proposed NA-MVP framework. Specifically, we investigate the impact of (1) bi-directional multi-view prompt learning, (2) UOT for local feature alignment, and (3) selective label refinement with OT and \(\phi_{i,k}\), as shown in Table~\ref{tab:ablation}.

\paragraph{Effect of bi-directional multi-view prompts}
To validate the role of bi-directional multi-view prompt learning, we compare the following variants: Our baseline is CoOp with one prompt. \textit{Negative Label}:  Employs one clean and one noisy prompt per class, while explicitly assigning a negative label to each image. \textit{Bi-directional}: Employs one clean and one noisy prompt per class, while treating non-target classes as implicit negatives. \textit{Multi-view}: Our complete design using multiple clean and noisy prompts. 
As shown in Table~\ref{tab:ablation} (a)-(d), employing both clean and noisy prompts already outperforms the single-prompt baseline, demonstrating the benefit of bi-directional supervision. However, the explicit negative-label strategy is less effective than our implicit negative design, as rigid counter-class assignments cannot generalize well under noisy conditions.
Further introducing multi-view prompts enhances robustness,  as it enriches semantic coverage and models intra-class diversity.

\begin{table}[t]
  \centering
  \caption{Test accuracy (\%) on Food101N.}
  \vspace{0.3em}  
  \label{tab:food}
  \setlength{\tabcolsep}{8pt}
  \begin{tabular}{@{}lccccc@{}}
    \toprule
    Method & 4-shot & 8-shot & 16-shot & 32-shot\\
    \midrule
    NLPrompt~\citep{pan2025nlprompt} & 70.57 & 73.93 & 76.46 & 76.87 \\
    NA-MVP & \textbf{76.10} & \textbf{76.27} & \textbf{76.90 }& \textbf{77.03}\\
    \bottomrule
  \end{tabular}
\end{table}

\begin{table}[t]
\label{tab:3}
  \centering
  \setlength{\tabcolsep}{4pt}  
  \caption{Ablation studies on DTD. (\%)}
  \label{tab:ablation}
  \begin{tabular}{@{}llcccc@{}}  
    \toprule
    & \textbf{Method/Noise Ratio} & \textbf{25\%} & \textbf{50\%} & \textbf{75\%} & \textbf{Avg} \\
    \midrule
    (a) & CoOp & 59.83 & 50.73 & 33.67 & 48.08 \\
    (b) & (a)+ Negative label& 59.53 & 52.53 & 34.40 & 48.82 \\
     (c) & (a)+ Bi-directional & 60.13 & 53.73 & 35.03 & 49.63 \\
     (d) & (c)+ Multi-view & 62.73 & 55.13 & 37.63 & 51.83 \\
    \midrule
    (e)
      & (d)+UOT  & 62.50 & 57.70 & 42.33 & 54.18 \\
      (f)& (d)+OT    & 62.30 & 56.80 & 41.00 & 53.37 \\
      (g)& (d)+KL Divergence & 62.27 & 56.43 & 39.10 & 52.60 \\
    \midrule
    (h)
      & (e)+OT refinement  & 59.60 & 54.77 & 45.77 & 53.38 \\
     (i) & (h)+$\phi_{i,k}$   & \textbf{63.13} & \textbf{58.50} & \textbf{48.63} & \textbf{56.75} \\
    \bottomrule
  \end{tabular}
\end{table}

\paragraph{Effect of UOT for patch-to-prompt alignment} To assess the impact of UOT for local feature alignment, we compare it with different distribution alignment methods, \textit{OT}: Enforces full mass matching between local features and prompts. \textit{KL Divergence}: Minimizes divergence between prompt and image patch distributions. \textit{UOT}: Introduces relaxed marginal constraints to focus on semantically relevant regions. As reported in Table~\ref{tab:ablation} (e)-(g), UOT achieves the best performance in all noise ratios. KL divergence struggles to handle multi-modal distributions from diverse prompts, and OT suffers from over-constrained mass transport, often forcing irrelevant alignments. In contrast, UOT provides the necessary flexibility to softly align multiple local features to diverse prompt semantics. In addition, we provide further comparisons with other matching methods like NBNN~\citep{boiman2008defense} in Appendix.
\begin{figure*}[t]
  \centering
  \begin{minipage}[t]{0.48\textwidth}
    \centering
    \includegraphics[width=\textwidth]{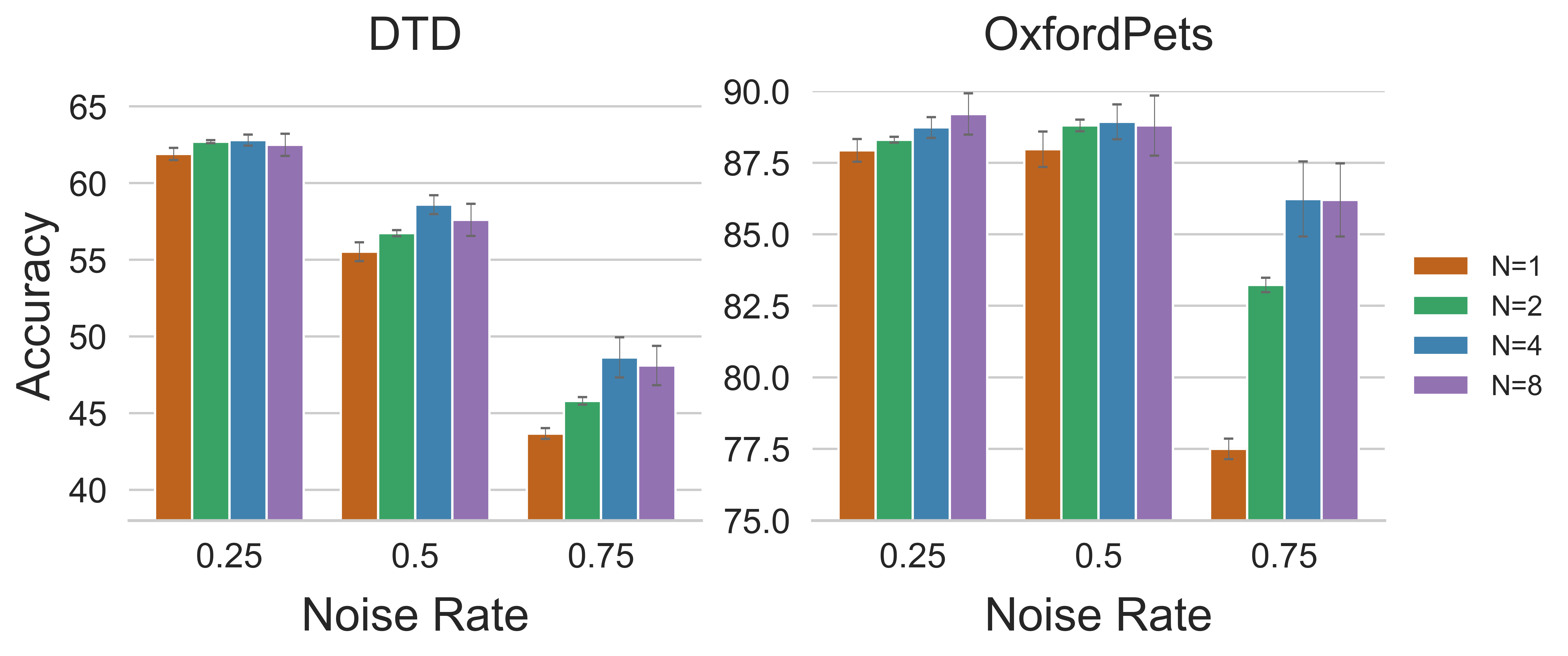}
    \caption{Test accuracy under varying label noise rates using different numbers of multi-view prompts $N \in \{1, 2, 4, 8\}$.}
    \label{fig:accuracy_noise}
  \end{minipage}
   \hfill
  \begin{minipage}[t]{0.48\textwidth}
    \centering
    \includegraphics[width=\textwidth]{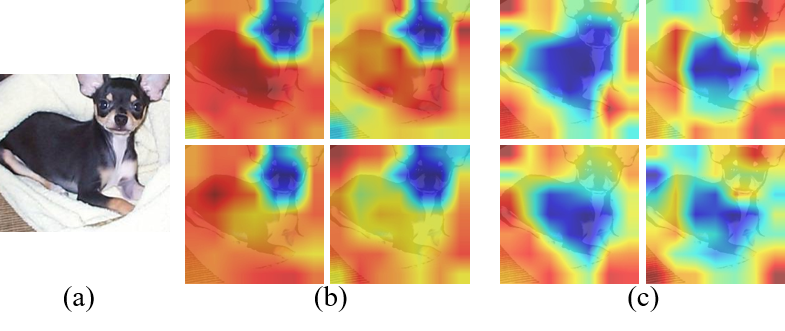}
    \caption{Visualization of bi-directional multi-view prompts. (a) The image; (b) The learned multi-view clean prompts; (c) The learned multi-view noisy prompts.}
    \label{fig:bi_view_prompt}
  \end{minipage} 
  \vspace{-5pt}
\end{figure*}

\paragraph{Effect of selective label refinement}
We investigate the effect of the proposed pseudo-label refinement strategy by comparing different variants of the global-level label correction mechanism: \textit{Refine by OT (w/o \(\phi_{i,k}\))}: this variant adopts a full label refinement strategy using OT, equivalent to using the label detection strategy used in NLPrompt and OT-Filter as noisy label identification strategy. \textit{Refine by OT (w/\(\phi_{i,k}\))}: partial refinement guided by \(\phi_{i,k}\). As shown in Table \ref{tab:ablation} (h)-(i), while full refinement improves performance under high noise levels, it can hurt accuracy when noise is low by mistakenly altering correct labels. In contrast, our selective refinement strategy (w/ \(\phi_{i,k}\)) consistently achieves better performance by focusing on likely noisy samples. 

To further investigate, we conducted further experiments on the OxfordPets dataset to compare our method with DEFT~\citep{wei2024vision}. DEFT selects clean samples using the rule: $p^{\text{c}}_{ik} > 0.5$. The results, summarized in Table \ref{tab:DEFT}, show that NA-MVP consistently outperforms DEFT across all noise levels, particularly under higher noise ratios. Unlike DEFT, which relies on rigid thresholds and employs coarse denoising strategies, NA-MVP’s selective refinement is more flexible, focusing on the most uncertain samples and effectively addressing noise. This selective approach prevents error propagation, ensuring more accurate and stable learning under noisy supervision. These findings highlight the advantages of NA-MVP in adapting to varying levels of noise, providing a more robust solution for noisy label problems. Additional analysis of selective label refinement quality is provided in the Appendix.

\begin{table}[t]
\centering
\setlength{\tabcolsep}{2.5pt}
\renewcommand{\arraystretch}{1}
\caption{The accuracy comparison on OxfordPets Dataset. }
\label{tab:DEFT}
\begin{tabular}{lcccccc}
\toprule
Noise rate & 12.5\% & 25\% & 37.5\% & 50\% & 62.5\% & 75\%  \\
\midrule
DEFT~\citep{wei2024vision}& \textbf{88.83} & 88.23 & 88.10 & 86.73 & 84.10 & 75.87 \\
NA-MVP            & 88.50 & \textbf{88.40} & \textbf{88.23} & \textbf{88.13} & \textbf{86.93} & \textbf{86.23} \\
\bottomrule
\vspace{-15pt}
\end{tabular}
\end{table}

\paragraph{Analysis of multi-view prompts}
To assess the effectiveness of learning multiple semantic views, we evaluate NA-MVP with different numbers of multi-view prompts ($N \in \{1, 2, 4, 8\}$) across various noise rates on DTD and OxfordPets. As shown in Figure~\ref{fig:accuracy_noise}, performance improves as $N$ increases from 1 to 4. This confirms our motivation that single-view prompt learning is insufficient for modeling the diverse and fine-grained cues needed to counteract noisy supervision in few-shot settings. Introducing multiple prompts allows the model to capture richer semantic patterns, improving both alignment accuracy and robustness to corrupted labels. However, further increasing to $N=8$ results in diminishing returns, suggesting that excessive prompts may cause redundancy. 
Thus, we adopt $N=4$ as the default, balancing robustness and efficiency. We also analyze the effect of imbalanced numbers of clean and noisy prompts in Appendix.

We also visualize the transport plan $T^*$ for four noisy and four clean prompts on the OxfordPets dataset in Figure~\ref{fig:bi_view_prompt}. The heatmaps reveal that noisy and clean prompts attend to different object attributes, highlighting that $T^*$ captures discriminative patterns helpful for noisy label learning. Visualizations of failure cases are also provided in Appendix.

\paragraph{Difference from prior frameworks}
Existing prompt-learning methods are typically designed for clean data supervised prompt–image alignment (PLOT~\citep{chen2022plot}), class-agnostic negative prompts (CLIPN~\citep{wang2023clipn}), or global OT relabeling
(NLPrompt~\citep{pan2025nlprompt}), they treat robustness as a \emph{global} alignment
problem. As a result, they cannot capture how clean and corrupted signals coexist and
interact within few-shot data, and their supervision remains sample-agnostic and brittle
to noise. NA-MVP introduces a fundamentally different view of robustness in prompt
learning for noisy labels: robustness should be achieved through \emph{sample-dependent,
fine-grained semantic alignment} rather than global matching. This perspective enables
NA-MVP to generate structured alignment signals that reveal the reliability of each
sample, forming the basis for more stable supervision under noisy few-shot conditions.

We provide additional materials in the appendix, including method details, comparisons with  other baselines, as well as hyper-parameter studies, generalization of NA-MVP, and computation cost evaluation.
\section{Conclusion}
We presented NA-MVP, a framework that rethinks robustness in prompt learning for noisy few-shot classification. Instead of relying on global image–prompt coupling or class-agnostic negative cues, NA-MVP is built on the idea that robustness should arise from sample-dependent, fine-grained semantic alignment. By combining bi-directional multi-view prompts with unbalanced patch-to-prompt transport, the framework produces structured alignment signals that naturally downweight corrupted regions and highlight stable semantics. These signals enable a selective OT refinement strategy that updates only unreliable samples, avoiding the over-correction of global pseudo-labeling methods. Experiments on synthetic and real-world noisy benchmarks demonstrate that NA-MVP consistently improves robustness over prior prompt-based and OT-based approaches in terms of robustness.

\section*{Acknowledgment}
This work was supported by the National Natural Science Foundation of
China under Grant 62401143, the State Key Project of Research and Development Plan under Grant 2024YFF1206703, the Natural Science Foundation of
Jiangsu Province under Grant BK20241301, and the Big Data Computing Center of Southeast University. AIIA refers to the Key Laboratory of New
Generation Artificial Intelligence Technology and Its Interdisciplinary Applications (Southeast University).
{
    \small
    \bibliographystyle{ieeenat_fullname}
    \bibliography{main}

\begin{thebibliography}{69}
\providecommand{\natexlab}[1]{#1}
\providecommand{\url}[1]{\texttt{#1}}
\expandafter\ifx\csname urlstyle\endcsname\relax
  \providecommand{\doi}[1]{doi: #1}\else
  \providecommand{\doi}{doi: \begingroup \urlstyle{rm}\Url}\fi

\bibitem[Arazo et~al.(2019)Arazo, Ortego, Albert, O’Connor, and McGuinness]{arazo2019unsupervised}
Eric Arazo, Diego Ortego, Paul Albert, Noel O’Connor, and Kevin McGuinness.
\newblock Unsupervised label noise modeling and loss correction.
\newblock In \emph{International conference on machine learning}, pages 312--321. PMLR, 2019.

\bibitem[Balaji et~al.(2020)Balaji, Chellappa, and Feizi]{balaji2020robust}
Yogesh Balaji, Rama Chellappa, and Soheil Feizi.
\newblock Robust optimal transport with applications in generative modeling and domain adaptation.
\newblock \emph{Advances in Neural Information Processing Systems}, 33:\penalty0 12934--12944, 2020.

\bibitem[Benamou et~al.(2015)Benamou, Carlier, Cuturi, Nenna, and Peyr{\'e}]{benamou2015iterative}
Jean-David Benamou, Guillaume Carlier, Marco Cuturi, Luca Nenna, and Gabriel Peyr{\'e}.
\newblock Iterative bregman projections for regularized transportation problems.
\newblock \emph{SIAM Journal on Scientific Computing}, 37\penalty0 (2):\penalty0 A1111--A1138, 2015.

\bibitem[Boiman et~al.(2008)Boiman, Shechtman, and Irani]{boiman2008defense}
Oren Boiman, Eli Shechtman, and Michal Irani.
\newblock In defense of nearest-neighbor based image classification.
\newblock In \emph{2008 IEEE conference on computer vision and pattern recognition}, pages 1--8. IEEE, 2008.

\bibitem[Chang et~al.(2017)Chang, Learned-Miller, and McCallum]{chang2017active}
Haw-Shiuan Chang, Erik Learned-Miller, and Andrew McCallum.
\newblock Active bias: Training more accurate neural networks by emphasizing high variance samples.
\newblock \emph{Advances in Neural Information Processing Systems}, 30, 2017.

\bibitem[Chang et~al.(2022)Chang, Shi, Tuan, and Wang]{chang2022unified}
Wanxing Chang, Ye Shi, Hoang Tuan, and Jingya Wang.
\newblock Unified optimal transport framework for universal domain adaptation.
\newblock \emph{Advances in Neural Information Processing Systems}, 35:\penalty0 29512--29524, 2022.

\bibitem[Chang et~al.(2023)Chang, Shi, and Wang]{chang2023csot}
Wanxing Chang, Ye Shi, and Jingya Wang.
\newblock Csot: Curriculum and structure-aware optimal transport for learning with noisy labels.
\newblock \emph{Advances in Neural Information Processing Systems}, 36:\penalty0 8528--8541, 2023.

\bibitem[Chen et~al.(2022)Chen, Yao, Song, Li, Rao, and Zhang]{chen2022plot}
Guangyi Chen, Weiran Yao, Xiangchen Song, Xinyue Li, Yongming Rao, and Kun Zhang.
\newblock Plot: Prompt learning with optimal transport for vision-language models.
\newblock \emph{arXiv preprint arXiv:2210.01253}, 2022.

\bibitem[Chizat et~al.(2018)Chizat, Peyr{\'e}, Schmitzer, and Vialard]{chizat2018unbalanced}
Lenaic Chizat, Gabriel Peyr{\'e}, Bernhard Schmitzer, and Fran{\c{c}}ois-Xavier Vialard.
\newblock Unbalanced optimal transport: Dynamic and kantorovich formulations.
\newblock \emph{Journal of Functional Analysis}, 274\penalty0 (11):\penalty0 3090--3123, 2018.

\bibitem[Choi et~al.(2023)Choi, Choi, and Kang]{choi2023generative}
Jaemoo Choi, Jaewoong Choi, and Myungjoo Kang.
\newblock Generative modeling through the semi-dual formulation of unbalanced optimal transport.
\newblock \emph{Advances in Neural Information Processing Systems}, 36:\penalty0 42433--42455, 2023.

\bibitem[Cimpoi et~al.(2014)Cimpoi, Maji, Kokkinos, Mohamed, and Vedaldi]{cimpoi2014describing}
Mircea Cimpoi, Subhransu Maji, Iasonas Kokkinos, Sammy Mohamed, and Andrea Vedaldi.
\newblock Describing textures in the wild.
\newblock In \emph{Proceedings of the IEEE conference on computer vision and pattern recognition}, pages 3606--3613, 2014.

\bibitem[Daniels et~al.(2021)Daniels, Maunu, and Hand]{daniels2021score}
Max Daniels, Tyler Maunu, and Paul Hand.
\newblock Score-based generative neural networks for large-scale optimal transport.
\newblock \emph{Advances in neural information processing systems}, 34:\penalty0 12955--12965, 2021.

\bibitem[De~Plaen et~al.(2023)De~Plaen, De~Plaen, Suykens, Proesmans, Tuytelaars, and Van~Gool]{de2023unbalanced}
Henri De~Plaen, Pierre-Fran{\c{c}}ois De~Plaen, Johan~AK Suykens, Marc Proesmans, Tinne Tuytelaars, and Luc Van~Gool.
\newblock Unbalanced optimal transport: A unified framework for object detection.
\newblock In \emph{Proceedings of the IEEE/CVF Conference on Computer Vision and Pattern Recognition}, pages 3198--3207, 2023.

\bibitem[Derakhshani et~al.(2023)Derakhshani, Sanchez, Bulat, da~Costa, Snoek, Tzimiropoulos, and Martinez]{derakhshani2023bayesian}
Mohammad~Mahdi Derakhshani, Enrique Sanchez, Adrian Bulat, Victor G~Turrisi da Costa, Cees~GM Snoek, Georgios Tzimiropoulos, and Brais Martinez.
\newblock Bayesian prompt learning for image-language model generalization.
\newblock In \emph{Proceedings of the IEEE/CVF International Conference on Computer Vision}, pages 15237--15246, 2023.

\bibitem[Distances(2013)]{distances2013lightspeed}
Cuturi M~Sinkhorn Distances.
\newblock Lightspeed computation of optimal transport.
\newblock \emph{Advances in neural information processing systems}, 26:\penalty0 2292--2300, 2013.

\bibitem[Fei-Fei et~al.(2004)Fei-Fei, Fergus, and Perona]{fei2004learning}
Li Fei-Fei, Rob Fergus, and Pietro Perona.
\newblock Learning generative visual models from few training examples: An incremental bayesian approach tested on 101 object categories.
\newblock In \emph{2004 conference on computer vision and pattern recognition workshop}, pages 178--178. IEEE, 2004.

\bibitem[Feng et~al.(2023)Feng, Ren, and Xie]{feng2023ot}
Chuanwen Feng, Yilong Ren, and Xike Xie.
\newblock Ot-filter: An optimal transport filter for learning with noisy labels.
\newblock In \emph{Proceedings of the IEEE/CVF Conference on Computer Vision and Pattern Recognition}, pages 16164--16174, 2023.

\bibitem[Frogner et~al.(2015)Frogner, Zhang, Mobahi, Araya, and Poggio]{frogner2015learning}
Charlie Frogner, Chiyuan Zhang, Hossein Mobahi, Mauricio Araya, and Tomaso~A Poggio.
\newblock Learning with a wasserstein loss.
\newblock \emph{Advances in neural information processing systems}, 28, 2015.

\bibitem[Ge et~al.(2021)Ge, Liu, Li, Yoshie, and Sun]{ge2021ota}
Zheng Ge, Songtao Liu, Zeming Li, Osamu Yoshie, and Jian Sun.
\newblock Ota: Optimal transport assignment for object detection.
\newblock In \emph{Proceedings of the IEEE/CVF conference on computer vision and pattern recognition}, pages 303--312, 2021.

\bibitem[Guo and Gu(2024)]{guo2024joapr}
Yuncheng Guo and Xiaodong Gu.
\newblock Joapr: Cleaning the lens of prompt learning for vision-language models.
\newblock In \emph{Proceedings of the IEEE/CVF Conference on Computer Vision and Pattern Recognition}, pages 28695--28705, 2024.

\bibitem[Iscen et~al.(2022)Iscen, Valmadre, Arnab, and Schmid]{iscen2022learning}
Ahmet Iscen, Jack Valmadre, Anurag Arnab, and Cordelia Schmid.
\newblock Learning with neighbor consistency for noisy labels.
\newblock In \emph{Proceedings of the IEEE/CVF conference on computer vision and pattern recognition}, pages 4672--4681, 2022.

\bibitem[Jia et~al.(2021)Jia, Yang, Xia, Chen, Parekh, Pham, Le, Sung, Li, and Duerig]{jia2021scaling}
Chao Jia, Yinfei Yang, Ye Xia, Yi-Ting Chen, Zarana Parekh, Hieu Pham, Quoc Le, Yun-Hsuan Sung, Zhen Li, and Tom Duerig.
\newblock Scaling up visual and vision-language representation learning with noisy text supervision.
\newblock In \emph{International conference on machine learning}, pages 4904--4916. PMLR, 2021.

\bibitem[Karim et~al.(2022)Karim, Rizve, Rahnavard, Mian, and Shah]{karim2022unicon}
Nazmul Karim, Mamshad~Nayeem Rizve, Nazanin Rahnavard, Ajmal Mian, and Mubarak Shah.
\newblock Unicon: Combating label noise through uniform selection and contrastive learning.
\newblock In \emph{Proceedings of the IEEE/CVF conference on computer vision and pattern recognition}, pages 9676--9686, 2022.

\bibitem[Khattak et~al.(2023{\natexlab{a}})Khattak, Rasheed, Maaz, Khan, and Khan]{khattak2023maple}
Muhammad~Uzair Khattak, Hanoona Rasheed, Muhammad Maaz, Salman Khan, and Fahad~Shahbaz Khan.
\newblock Maple: Multi-modal prompt learning.
\newblock In \emph{Proceedings of the IEEE/CVF conference on computer vision and pattern recognition}, pages 19113--19122, 2023{\natexlab{a}}.

\bibitem[Khattak et~al.(2023{\natexlab{b}})Khattak, Wasim, Naseer, Khan, Yang, and Khan]{khattak2023self}
Muhammad~Uzair Khattak, Syed~Talal Wasim, Muzammal Naseer, Salman Khan, Ming-Hsuan Yang, and Fahad~Shahbaz Khan.
\newblock Self-regulating prompts: Foundational model adaptation without forgetting.
\newblock In \emph{Proceedings of the IEEE/CVF international conference on computer vision}, pages 15190--15200, 2023{\natexlab{b}}.

\bibitem[Kim et~al.(2021)Kim, Ko, Choi, Yun, et~al.]{kim2021fine}
Taehyeon Kim, Jongwoo Ko, JinHwan Choi, Se-Young Yun, et~al.
\newblock Fine samples for learning with noisy labels.
\newblock \emph{Advances in Neural Information Processing Systems}, 34:\penalty0 24137--24149, 2021.

\bibitem[Ko et~al.(2023)Ko, Yi, and Yun]{ko2023gift}
Jongwoo Ko, Bongsoo Yi, and Se-Young Yun.
\newblock A gift from label smoothing: robust training with adaptive label smoothing via auxiliary classifier under label noise.
\newblock In \emph{Proceedings of the AAAI Conference on Artificial Intelligence}, pages 8325--8333, 2023.

\bibitem[Lahn et~al.(2023)Lahn, Raghvendra, and Zhang]{lahn2023combinatorial}
Nathaniel Lahn, Sharath Raghvendra, and Kaiyi Zhang.
\newblock A combinatorial algorithm for approximating the optimal transport in the parallel and mpc settings.
\newblock \emph{Advances in Neural Information Processing Systems}, 36:\penalty0 21675--21686, 2023.

\bibitem[Lai et~al.(2022)Lai, Wang, Cheung, and Chuah]{lai2022sar}
Zhengfeng Lai, Chao Wang, Sen-ching Cheung, and Chen-Nee Chuah.
\newblock Sar: Self-adaptive refinement on pseudo labels for multiclass-imbalanced semi-supervised learning.
\newblock In \emph{Proceedings of the IEEE/CVF Conference on Computer Vision and Pattern Recognition}, pages 4091--4100, 2022.

\bibitem[Lee et~al.(2018)Lee, He, Zhang, and Yang]{lee2018cleannet}
Kuang-Huei Lee, Xiaodong He, Lei Zhang, and Linjun Yang.
\newblock Cleannet: Transfer learning for scalable image classifier training with label noise.
\newblock In \emph{Proceedings of the IEEE conference on computer vision and pattern recognition}, pages 5447--5456, 2018.

\bibitem[Li et~al.(2023)Li, Han, Shan, and Chen]{li2023disc}
Yifan Li, Hu Han, Shiguang Shan, and Xilin Chen.
\newblock Disc: Learning from noisy labels via dynamic instance-specific selection and correction.
\newblock In \emph{Proceedings of the IEEE/CVF conference on computer vision and pattern recognition}, pages 24070--24079, 2023.

\bibitem[Liu et~al.(2023)Liu, Lu, Liu, An, Xu, Yao, Zhang, Xiong, and Gui]{liu2023hierarchical}
Yajing Liu, Yuning Lu, Hao Liu, Yaozu An, Zhuoran Xu, Zhuokun Yao, Baofeng Zhang, Zhiwei Xiong, and Chenguang Gui.
\newblock Hierarchical prompt learning for multi-task learning.
\newblock In \emph{Proceedings of the IEEE/CVF Conference on Computer Vision and Pattern Recognition}, pages 10888--10898, 2023.

\bibitem[Lombardi and Maitre(2015)]{lombardi2015eulerian}
Damiano Lombardi and Emmanuel Maitre.
\newblock Eulerian models and algorithms for unbalanced optimal transport.
\newblock \emph{ESAIM: Mathematical Modelling and Numerical Analysis}, 49\penalty0 (6):\penalty0 1717--1744, 2015.

\bibitem[Lu and He(2022)]{lu2022selc}
Yangdi Lu and Wenbo He.
\newblock Selc: self-ensemble label correction improves learning with noisy labels.
\newblock \emph{arXiv preprint arXiv:2205.01156}, 2022.

\bibitem[Lu et~al.(2022)Lu, Liu, Zhang, Liu, and Tian]{lu2022prompt}
Yuning Lu, Jianzhuang Liu, Yonggang Zhang, Yajing Liu, and Xinmei Tian.
\newblock Prompt distribution learning.
\newblock In \emph{Proceedings of the IEEE/CVF Conference on Computer Vision and Pattern Recognition}, pages 5206--5215, 2022.

\bibitem[Ma et~al.(2020)Ma, Huang, Wang, Romano, Erfani, and Bailey]{ma2020normalized}
Xingjun Ma, Hanxun Huang, Yisen Wang, Simone Romano, Sarah Erfani, and James Bailey.
\newblock Normalized loss functions for deep learning with noisy labels.
\newblock In \emph{International conference on machine learning}, pages 6543--6553. PMLR, 2020.

\bibitem[Nilsback and Zisserman(2008)]{nilsback2008automated}
Maria-Elena Nilsback and Andrew Zisserman.
\newblock Automated flower classification over a large number of classes.
\newblock In \emph{2008 Sixth Indian conference on computer vision, graphics \& image processing}, pages 722--729. IEEE, 2008.

\bibitem[Pan et~al.(2025)Pan, Li, Tang, Huang, Fang, Liu, Wang, Yu, and Shi]{pan2025nlprompt}
Bikang Pan, Qun Li, Xiaoying Tang, Wei Huang, Zhen Fang, Feng Liu, Jingya Wang, Jingyi Yu, and Ye Shi.
\newblock Nlprompt: Noise-label prompt learning for vision-language models.
\newblock In \emph{Proceedings of the Computer Vision and Pattern Recognition Conference}, pages 19963--19973, 2025.

\bibitem[Parkhi et~al.(2012)Parkhi, Vedaldi, Zisserman, and Jawahar]{parkhi2012cats}
Omkar~M Parkhi, Andrea Vedaldi, Andrew Zisserman, and CV Jawahar.
\newblock Cats and dogs.
\newblock In \emph{2012 IEEE conference on computer vision and pattern recognition}, pages 3498--3505. IEEE, 2012.

\bibitem[Peng et~al.(2018)Peng, Zhao, and Zhang]{peng2018two}
Yuxin Peng, Yunzhen Zhao, and Junchao Zhang.
\newblock Two-stream collaborative learning with spatial-temporal attention for video classification.
\newblock \emph{IEEE Transactions on Circuits and Systems for Video Technology}, 29\penalty0 (3):\penalty0 773--786, 2018.

\bibitem[Peyr{\'e} et~al.(2019)Peyr{\'e}, Cuturi, et~al.]{peyre2019computational}
Gabriel Peyr{\'e}, Marco Cuturi, et~al.
\newblock Computational optimal transport: With applications to data science.
\newblock \emph{Foundations and Trends{\textregistered} in Machine Learning}, 11\penalty0 (5-6):\penalty0 355--607, 2019.

\bibitem[Radford et~al.(2021)Radford, Kim, Hallacy, Ramesh, Goh, Agarwal, Sastry, Askell, Mishkin, Clark, et~al.]{radford2021learning}
Alec Radford, Jong~Wook Kim, Chris Hallacy, Aditya Ramesh, Gabriel Goh, Sandhini Agarwal, Girish Sastry, Amanda Askell, Pamela Mishkin, Jack Clark, et~al.
\newblock Learning transferable visual models from natural language supervision.
\newblock In \emph{International conference on machine learning}, pages 8748--8763. PMLR, 2021.

\bibitem[Roy and Etemad(2023)]{roy2023consistency}
Shuvendu Roy and Ali Etemad.
\newblock Consistency-guided prompt learning for vision-language models.
\newblock \emph{arXiv preprint arXiv:2306.01195}, 2023.

\bibitem[Shu et~al.(2022)Shu, Nie, Huang, Yu, Goldstein, Anandkumar, and Xiao]{shu2022test}
Manli Shu, Weili Nie, De-An Huang, Zhiding Yu, Tom Goldstein, Anima Anandkumar, and Chaowei Xiao.
\newblock Test-time prompt tuning for zero-shot generalization in vision-language models.
\newblock \emph{Advances in Neural Information Processing Systems}, 35:\penalty0 14274--14289, 2022.

\bibitem[Sun et~al.(2022)Sun, Hu, and Saenko]{sun2022dualcoop}
Ximeng Sun, Ping Hu, and Kate Saenko.
\newblock Dualcoop: Fast adaptation to multi-label recognition with limited annotations.
\newblock \emph{Advances in Neural Information Processing Systems}, 35:\penalty0 30569--30582, 2022.

\bibitem[Taherkhani et~al.(2020)Taherkhani, Dabouei, Soleymani, Dawson, and Nasrabadi]{taherkhani2020transporting}
Fariborz Taherkhani, Ali Dabouei, Sobhan Soleymani, Jeremy Dawson, and Nasser~M Nasrabadi.
\newblock Transporting labels via hierarchical optimal transport for semi-supervised learning.
\newblock In \emph{Computer Vision--ECCV 2020: 16th European Conference, Glasgow, UK, August 23--28, 2020, Proceedings, Part IV 16}, pages 509--526. Springer, 2020.

\bibitem[Wang et~al.(2022)Wang, Xia, Li, Mao, Feng, Chen, and Zhao]{wang2022solar}
Haobo Wang, Mingxuan Xia, Yixuan Li, Yuren Mao, Lei Feng, Gang Chen, and Junbo Zhao.
\newblock Solar: Sinkhorn label refinery for imbalanced partial-label learning.
\newblock \emph{Advances in neural information processing systems}, 35:\penalty0 8104--8117, 2022.

\bibitem[Wang et~al.(2023)Wang, Li, Yao, and Li]{wang2023clipn}
Hualiang Wang, Yi Li, Huifeng Yao, and Xiaomeng Li.
\newblock Clipn for zero-shot ood detection: Teaching clip to say no.
\newblock In \emph{Proceedings of the IEEE/CVF International Conference on Computer Vision}, pages 1802--1812, 2023.

\bibitem[Wang et~al.(2019)Wang, Ma, Chen, Luo, Yi, and Bailey]{wang2019symmetric}
Yisen Wang, Xingjun Ma, Zaiyi Chen, Yuan Luo, Jinfeng Yi, and James Bailey.
\newblock Symmetric cross entropy for robust learning with noisy labels.
\newblock In \emph{Proceedings of the IEEE/CVF international conference on computer vision}, pages 322--330, 2019.

\bibitem[Wang et~al.(2024)Wang, Shen, Zavlanos, and Johansson]{wang2024outlier}
Zifan Wang, Yi Shen, Michael Zavlanos, and Karl~H Johansson.
\newblock Outlier-robust distributionally robust optimization via unbalanced optimal transport.
\newblock \emph{Advances in Neural Information Processing Systems}, 37:\penalty0 52189--52214, 2024.

\bibitem[Wei et~al.(2020)Wei, Feng, Chen, and An]{wei2020combating}
Hongxin Wei, Lei Feng, Xiangyu Chen, and Bo An.
\newblock Combating noisy labels by agreement: A joint training method with co-regularization.
\newblock In \emph{Proceedings of the IEEE/CVF conference on computer vision and pattern recognition}, pages 13726--13735, 2020.

\bibitem[Wei et~al.(2021)Wei, Tao, Xie, and An]{wei2021open}
Hongxin Wei, Lue Tao, Renchunzi Xie, and Bo An.
\newblock Open-set label noise can improve robustness against inherent label noise.
\newblock \emph{Advances in Neural Information Processing Systems}, 34:\penalty0 7978--7992, 2021.

\bibitem[Wei et~al.(2023)Wei, Zhuang, Xie, Feng, Niu, An, and Li]{wei2023mitigating}
Hongxin Wei, Huiping Zhuang, Renchunzi Xie, Lei Feng, Gang Niu, Bo An, and Yixuan Li.
\newblock Mitigating memorization of noisy labels by clipping the model prediction.
\newblock In \emph{International conference on machine learning}, pages 36868--36886. PMLR, 2023.

\bibitem[Wei et~al.(2024)Wei, Li, Li, Shi, Li, and Zhang]{wei2024vision}
Tong Wei, Hao-Tian Li, ChunShu Li, Jiang-Xin Shi, Yu-Feng Li, and Min-Ling Zhang.
\newblock Vision-language models are strong noisy label detectors.
\newblock \emph{Advances in Neural Information Processing Systems}, 37:\penalty0 58154--58173, 2024.

\bibitem[Wu et~al.(2023)Wu, Tian, Yu, Wang, Morgado, Hu, and Yang]{wu2023prompt}
Cheng-En Wu, Yu Tian, Haichao Yu, Heng Wang, Pedro Morgado, Yu~Hen Hu, and Linjie Yang.
\newblock Why is prompt tuning for vision-language models robust to noisy labels?
\newblock In \emph{Proceedings of the IEEE/CVF International Conference on Computer Vision}, pages 15488--15497, 2023.

\bibitem[Xia et~al.(2022)Xia, Tan, Wu, Xu, and Li]{xia2022ot}
Jun Xia, Cheng Tan, Lirong Wu, Yongjie Xu, and Stan~Z Li.
\newblock Ot cleaner: Label correction as optimal transport.
\newblock In \emph{ICASSP 2022-2022 IEEE International Conference on Acoustics, Speech and Signal Processing (ICASSP)}, pages 3953--3957. IEEE, 2022.

\bibitem[Xia et~al.(2019)Xia, Liu, Wang, Han, Gong, Niu, and Sugiyama]{xia2019anchor}
Xiaobo Xia, Tongliang Liu, Nannan Wang, Bo Han, Chen Gong, Gang Niu, and Masashi Sugiyama.
\newblock Are anchor points really indispensable in label-noise learning?
\newblock \emph{Advances in neural information processing systems}, 32, 2019.

\bibitem[Xu et~al.(2023)Xu, Niu, Yang, Drew, Zhou, and Chen]{xu2023usdnl}
Yuanzhuo Xu, Xiaoguang Niu, Jie Yang, Steve Drew, Jiayu Zhou, and Ruizhi Chen.
\newblock Usdnl: Uncertainty-based single dropout in noisy label learning.
\newblock In \emph{Proceedings of the AAAI Conference on Artificial Intelligence}, pages 10648--10656, 2023.

\bibitem[Xue(2025)]{xue2025dier}
Cheng Xue.
\newblock Dier-net: Debiased learning with medical image noisy label by intrinsic and extrinsic regularization.
\newblock \emph{International Journal of Imaging Systems and Technology}, 35\penalty0 (4):\penalty0 e70160, 2025.

\bibitem[Xue et~al.(2019)Xue, Dou, Shi, Chen, and Heng]{xue2019robust}
Cheng Xue, Qi Dou, Xueying Shi, Hao Chen, and Pheng-Ann Heng.
\newblock Robust learning at noisy labeled medical images: Applied to skin lesion classification.
\newblock In \emph{2019 IEEE 16th International symposium on biomedical imaging (ISBI 2019)}, pages 1280--1283. IEEE, 2019.

\bibitem[Xue et~al.(2020)Xue, Deng, Li, Dou, and Heng]{xue2020cascaded}
Cheng Xue, Qiao Deng, Xiaomeng Li, Qi Dou, and Pheng-Ann Heng.
\newblock Cascaded robust learning at imperfect labels for chest x-ray segmentation.
\newblock In \emph{International conference on medical image computing and computer-assisted intervention}, pages 579--588. Springer, 2020.

\bibitem[Xue et~al.(2022)Xue, Yu, Chen, Dou, and Heng]{xue2022robust}
Cheng Xue, Lequan Yu, Pengfei Chen, Qi Dou, and Pheng-Ann Heng.
\newblock Robust medical image classification from noisy labeled data with global and local representation guided co-training.
\newblock \emph{IEEE transactions on medical imaging}, 41\penalty0 (6):\penalty0 1371--1382, 2022.

\bibitem[Yang et~al.(2021)Yang, Yan, Ming, Wang, Zhang, and Tian]{yang2021rethinking}
Xue Yang, Junchi Yan, Qi Ming, Wentao Wang, Xiaopeng Zhang, and Qi Tian.
\newblock Rethinking rotated object detection with gaussian wasserstein distance loss.
\newblock In \emph{International conference on machine learning}, pages 11830--11841. PMLR, 2021.

\bibitem[Yu et~al.(2022)Yu, Wang, Vasudevan, Yeung, Seyedhosseini, and Wu]{yu2022coca}
Jiahui Yu, Zirui Wang, Vijay Vasudevan, Legg Yeung, Mojtaba Seyedhosseini, and Yonghui Wu.
\newblock Coca: Contrastive captioners are image-text foundation models.
\newblock \emph{arXiv preprint arXiv:2205.01917}, 2022.

\bibitem[Zhang and Sabuncu(2018)]{zhang2018generalized}
Zhilu Zhang and Mert Sabuncu.
\newblock Generalized cross entropy loss for training deep neural networks with noisy labels.
\newblock \emph{Advances in neural information processing systems}, 31, 2018.

\bibitem[Zheng et~al.(2021)Zheng, Awadallah, and Dumais]{zheng2021meta}
Guoqing Zheng, Ahmed~Hassan Awadallah, and Susan Dumais.
\newblock Meta label correction for noisy label learning.
\newblock In \emph{Proceedings of the AAAI conference on artificial intelligence}, pages 11053--11061, 2021.

\bibitem[Zhou et~al.(2022{\natexlab{a}})Zhou, Yang, Loy, and Liu]{zhou2022conditional}
Kaiyang Zhou, Jingkang Yang, Chen~Change Loy, and Ziwei Liu.
\newblock Conditional prompt learning for vision-language models.
\newblock In \emph{Proceedings of the IEEE/CVF conference on computer vision and pattern recognition}, pages 16816--16825, 2022{\natexlab{a}}.

\bibitem[Zhou et~al.(2022{\natexlab{b}})Zhou, Yang, Loy, and Liu]{zhou2022learning}
Kaiyang Zhou, Jingkang Yang, Chen~Change Loy, and Ziwei Liu.
\newblock Learning to prompt for vision-language models.
\newblock \emph{International Journal of Computer Vision}, 130\penalty0 (9):\penalty0 2337--2348, 2022{\natexlab{b}}.

\bibitem[Zhu et~al.(2023)Zhu, Niu, Han, Wu, and Zhang]{zhu2023prompt}
Beier Zhu, Yulei Niu, Yucheng Han, Yue Wu, and Hanwang Zhang.
\newblock Prompt-aligned gradient for prompt tuning.
\newblock In \emph{Proceedings of the IEEE/CVF international conference on computer vision}, pages 15659--15669, 2023.

\end{thebibliography}
}


\end{document}